\documentclass{article}

\usepackage{aaai}

\usepackage{graphicx}
\usepackage{color}
\usepackage[fleqn]{amsmath}
\usepackage{amssymb}
\usepackage[all]{xy}
\usepackage{float}
\usepackage{xspace}
\usepackage{url}
\usepackage[compact]{titlesec}

\floatstyle{boxed}
\restylefloat{figure}

\newfloat{program}{thp}{}
\floatname{program}{Program}

\newcommand{\comment}[1]{{}}

\newcommand{\myc}[1]{}

\newtheorem{Algo}{Algorithm}
\newtheorem{Lem}{Lemma}

\newtheorem{Thm}[Lem]{Theorem}
\newtheorem{Pro}[Lem]{Proposition}

\newtheorem{Def}{Definition}

\newtheorem{Ex}{Example}

\newcommand{\id}[1]{\mbox{\it #1\/}}

\newcommand{\kw}[1]{\mbox{\tt #1}}

\def\p@enumiii{\theenumi(\theenumii)}

\renewcommand{\subparagraph}[1]{\smallskip \noindent \textbf{\textit{#1}} \hspace*{0.5em}}
\newcommand{\mgu}{\mbox{\rm mgu\xspace}}
\newcommand{\rv}{\mbox{\rm rv\xspace}}

\def\squareforqed{\hbox{\rlap{$\sqcap$}$\sqcup$}}
\def\qed{\ifmmode\squareforqed\else{\unskip\nobreak\hfil
\penalty50\hskip1em\null\nobreak\hfil\squareforqed
\parfillskip=0pt\finalhyphendemerits=0\endgraf}\fi}

\newcommand\annotate[1]%
{\textcolor{blue}{$\maltese$} %
\-\marginpar[\raggedleft\scriptsize \textcolor{red}{#1}]%
{\raggedright\scriptsize \textcolor{red}{#1}}}

\newcommand{\mycite}[1]{\mbox{\citeauthor{#1} (\citeyear{#1})}}

\title{Parameter Learning in PRISM~Programs with Continuous~Random~Variables}

\author{ {\bf Muhammad Asiful Islam, C.R. Ramakrishnan, I.V. Ramakrishnan} \\
Dept. of Computer Science\\
Stony Brook University\\
Stony Brook, NY 11794 \\
\{maislam, cram, ram\}@cs.sunysb.edu\\
}

\begin{document}

\maketitle

\begin{abstract}
  Probabilistic Logic Programming (PLP), exemplified by Sato and
  Kameya's PRISM, Poole's ICL, De Raedt et al's ProbLog and Vennekens et
  al's LPAD, combines statistical and logical knowledge representation
  and inference.  Inference in these languages is based on
  enumerative construction of proofs over logic programs.
  Consequently, these languages permit very limited use of random
  variables with continuous distributions.  In this paper, we extend
  PRISM with Gaussian random variables and linear equality
  constraints, and consider the problem of parameter learning in the
  extended language.  Many statistical models such as finite mixture
  models and Kalman filter can be encoded in extended PRISM.  Our
  EM-based learning algorithm uses a symbolic inference procedure that
  represents sets of derivations without enumeration.  This permits us
  to learn the distribution parameters of extended PRISM programs with
  discrete as well as Gaussian variables.  The learning algorithm
  naturally generalizes the ones used for PRISM and Hybrid Bayesian
  Networks.
\end{abstract}

%%% Local Variables:
%%% mode: latex
%%% TeX-master: "main"
%%% End:

\section{Introduction}
\label{sec:intro}

Probabilistic Logic Programming (PLP) is a 
class of Statistical Relational Learning
(SRL) frameworks~\cite{srlbook} which combine
statistical and logical knowledge representation and inference.
PLP languages, such as SLP~\cite{Muggleton}, ICL~\cite{PooleICL},
PRISM~\cite{sato-kameya-prism}, ProbLog~\cite{deRaedt} and
LPAD~\cite{lpad}  extend traditional logic
programming languages by implicitly or explicitly attaching 
random variables with certain clauses in a logic
program.  A large class of common statistical models, such as Bayesian
networks, Hidden Markov models and Probabilistic Context-Free Grammars
have been effectively encoded in PLP; the programming aspect of PLP
has also been exploited to succinctly specify complex models, such as
discovering links in biological networks~\cite{deRaedt}.  Parameter learning in these
languages is typically done by variants of the EM
algorithm~\cite{DempsterEM}.  

Operationally, combined statistical/logical inference in PLP is
based on proof structures similar to those created by
pure logical inference.  As a result, these languages have 
limited support models with continuous random
variables.  Recently, we extended PRISM~\cite{sato-kameya-prism}
with Gaussian and Gamma-distributed random variables, and
linear equality constraints (\url{http://arxiv.org/abs/1112.2681})\footnote{Relevant
  technical aspects of this extension are summarized in this paper to make it self-contained.}. This extension permits encoding of
complex statistical models including Kalman filters and a large class
of Hybrid Bayesian Networks.  

In this paper, we present an algorithm for parameter learning in PRISM
extended with Gaussian random variables.  The key aspect of this
algorithm is the construction of \emph{symbolic derivations} that
succinctly represent large (sometimes infinite) sets of traditional
logical derivations.  Our learning algorithm represents and computes
Expected Sufficient Statistics (ESS) symbolically as well, for
Gaussian as well as discrete random variables. 
Although our technical development is limited to PRISM, the core algorithm can be
adapted to parameter learning in (extended versions of) other PLP
languages as well.

\paragraph{Related Work}
SRL frameworks can be broadly classified as
statistical-model-based or logic-based, depending on how their
semantics is defined.  In the first category are languages such as
Bayesian Logic Programs (BLPs)~\cite{hblp}, Probabilistic Relational
Models (PRMs)~\cite{hprm}, and Markov Logic Networks (MLNs)~\cite{mln},
where logical relations are used to specify a model compactly.
Although originally defined over discrete random variables, these
languages have been extended (e.g.  Continuous
BLP~\cite{hblp}, Hybrid PRM~\cite{hprm}, and Hybrid
MLN~\cite{hmln}) to support continuous random variables as
well.  Techniques for parameter learning in statistical-model-based
languages are adapted from the corresponding techniques in the
underlying statistical models.  For example, discriminative learning
techniques are used for parameter learning in MLNs
~\cite{mlnlearningA,mlnlearningB}.

Logic-based SRL languages include the PLP languages mentioned earlier.
Hybrid ProbLog ~\cite{hproblog} extends ProbLog by adding
continuous probabilistic facts, but restricts their use such that
statistical models such as Kalman filters and certain classes of Hybrid Bayesian Networks
(with continuous child with continuous parents) 
cannot be encoded.  More recently, \mycite{apprProblog} introduced a
sampling-based approach for (approximate) probabilistic inference in a
ProbLog-like language. 

Graphical EM~\cite{sato} is the parameter learning algorithm used in
PRISM.  Interestingly, graphical EM reduces to the
Baum-Welch~\cite{rabiner} algorithm for HMMs encoded in PRISM.
\mycite{probloglearningA} introduced a least squares optimization
approach to learn distribution parameters in ProbLog.
CoPrEM~\cite{probloglearningB} is another algorithm for ProLog that
computes binary decision diagrams (BDDs) for representing proofs and
uses a dynamic programming approach to estimate
parameters. BO-EM~\cite{satoBDD} is a BDD-based parameter learning
algorithm for PRISM.  These techniques enumerate derivations (even
when represented as BDDs), and do not readily generalize when
continuous random variables are introduced.

\section{Background: An Overview of PRISM}
\label{sec:prism}

%PAGE BUDGET: 1.0 page including Extended PRISM.

PRISM programs have Prolog-like syntax (see Example~\ref{ex:fmm}).
In a PRISM program the \texttt{msw} relation (``multi-valued switch'')
has a special meaning: \texttt{msw(X,I,V)} says that \texttt{V} is a
random variable.  More precisely, \texttt{V} is the outcome of the
\texttt{I}-th instance from a family \texttt{X} of random
processes\footnote{Following PRISM, we often omit the instance number
  in an \texttt{msw} when a program uses only one instance from a
  family of random processes.}.  The set of variables $\{\mathtt{V}_i
\mid \mathtt{msw(}p, i, \mathtt{V}_i\mathtt{)}\}$ are i.i.d., and
their distribution is given by the random process $p$.  The
\texttt{msw} relation provides the mechanism for using random
variables, thereby allowing us to weave together statistical and
logical aspects of a model into a single program.  The distribution
parameters of the random variables are specified separately.

PRISM programs have declarative semantics, called 
\emph{distribution semantics}~\cite{sato-kameya-prism,sato}.
Operationally,  query evaluation in PRISM closely follows that for
traditional logic programming, with one
modification.  When the goal selected at a step is of the form
\texttt{msw(X,I,Y)}, then \texttt{Y} is bound to a possible outcome of a
random process \texttt{X}.  The derivation step is
associated with the probability of this outcome.  If all random
processes encountered in a derivation are independent, then the
probability of the derivation is the product of probabilities of each
step in the derivation.  If a set of derivations are pairwise mutually
exclusive, the probability of the set is the sum of probabilities of
each derivation in the set\footnote{The evaluation procedure is
  defined only when the independence and exclusiveness assumptions
  hold.}.  
Finally, the probability of an answer to
a query is computed as the probability of the set of derivations
corresponding to that answer.  

As an illustration, consider the query \texttt{fmix(X)} evaluated over program in
Example~\ref{ex:fmm}.   One step of resolution derives goal of
the form \texttt{msw(m, M), msw(w(M),X)}. Now depending on the value of \texttt{m},
there are two possible next steps:
\texttt{msw(w(a),X)} and \texttt{msw(w(b),X)}.
\emph{Thus in PRISM, derivations are constructed by enumerating the
  possible outcomes of each random variable.}

\begin{Ex}[Finite Mixture Model]
In the following PRISM program, which encodes a finite mixture model~\cite{fmm},
{\rm \texttt{msw(m, M)}} chooses one distribution from a finite set of
continuous 
distributions, {\rm \texttt{msw(w(M), X)}} samples {\rm \texttt{X}} from the chosen distribution.

\begin{small}
  \begin{minipage}{3.0in}
\begin{verbatim}
fmix(X) :- msw(m, M),
           msw(w(M), X).

% Ranges of RVs
values(m, [a,b]).
values(w(M), real).
% PDFs and PMFs
:- set_sw(m, [0.3, 0.7]),
   set_sw(w(a), norm(2.0, 1.0)),
   set_sw(w(b), norm(3.0, 1.0)).
\end{verbatim}
\end{minipage}
\end{small}
\qed
  \label{ex:fmm}
\end{Ex}

\section{Extended PRISM}
\label{sec:language}

Support for continuous variables is added by modifying PRISM's
language in two ways.   We use the \texttt{msw}
relation to sample from discrete as well as continuous distributions.
In PRISM, a special relation
called \texttt{values} is used to specify the ranges of values of
random variables; the probability mass functions are specified using
\texttt{set\_sw} directives.  We extend the
\texttt{set\_sw} directives to specify probability density functions
as well.  For instance, \texttt{set\_sw(r, norm(Mu,Var))} specifies
that outcomes of  random processes \texttt{r} have Gaussian
distribution with mean 
\texttt{Mu} and variance \texttt{Var}.
\comment{ \footnote{The technical
  development in this paper considers only univariate Gaussian
  variables; see Discussions section on a discussion on how
  multivariate Gaussian as well as other continuous distributions are
  handled.}}  Parameterized families of random
processes may be specified, as long as the parameters are
discrete-valued.  For 
instance, \texttt{set\_sw(w(M), norm(Mu,Var))} specifies a family of
random processes, with one for each value of \texttt{M}. As
in PRISM, \texttt{set\_sw} directives may be specified
programmatically; for instance, the distribution parameters of \texttt{w(M)},
 may be computed as functions of
\texttt{M}.

Additionally, we extend PRISM programs with linear equality
constraints over reals.  Without loss of generality, we
assume that constraints are written as linear equalities of the form
$Y = a_1 * X_1 + \ldots + a_n * X_n + b$ where $a_i$ and $b$ are all
floating-point constants.
\comment{ , or inequalities of the form $Y < a$ or $Y>a$
for some floating point constant $a$. 
Note that inequalities comparing
two variables can be expressed as an inequality comparing a linear
function of the two variables and a constant.}
The use of constraints enables us to encode Hybrid Bayesian Networks and Kalman Filters as
extended PRISM programs.  In the following, we use \emph{Constr} to
denote a set (conjunction) of linear equality 
constraints.  We also denote by $\overline{X}$ a vector of variables
and/or values, explicitly specifying the size only when it is not
clear from the context.  This permits us to write linear equality
constraints compactly (e.g., $Y = \overline{a}\cdot\overline{X} +b$).

%%% Local Variables:
%%% mode: latex
%%% TeX-master: "main"
%%% End:

\section{Inference}
\label{sec:inference}

\newcommand{\marginalize}{\mathbb{M}}
\newcommand{\integrate}{\mathbb{I}}
\newcommand{\project}{\mathbb{P}}

The key to inference in the presence of
continuous random variables is avoiding enumeration by 
representing the derivations and their attributes symbolically.  A single
step in the construction of a symbolic derivation is defined below.

\begin{figure}
\centering
\begin{minipage}[t]{1.5in}
\begin{displaymath}
\scriptsize{
  \xymatrix @=10pt {
    G_{1}: fmix(X) \ar[d]  \\
    G_{2}: msw(m,M), msw(w(M), X).  \ar[d]   \\
    G_{3}: msw(w(M), X).  
  }
}
\end{displaymath}
\end{minipage}
\caption{Symbolic derivation for goal \texttt{fmix(X)}}
\label{fig:fmmderivation}
\end{figure}
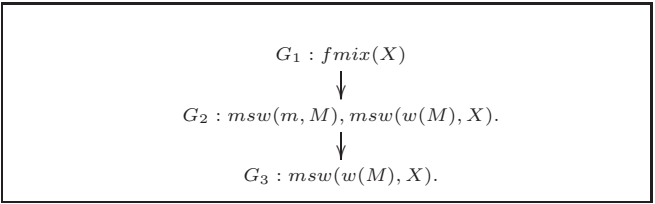

\begin{Def}[Symbolic Derivation]
A goal $G$ \emph{directly derives}  goal $G'$, denoted $G \rightarrow
G'$,  if:
\begin{description}
\item[PCR:] $G = q_1(\overline{X_1}), G_1$, and there exists a
  clause in the program, $q_1(\overline{Y}) \mbox{:-} r_1(\overline{Y_1}),
  r_2(\overline{Y_2}), \ldots, r_m(\overline{Y_m})$, such that $\theta =
  \mgu(q_1(\overline{X_1}), q_1(\overline{Y}))$; then, $G' =
  (r_1(\overline{Y_1}), r_2(\overline{Y_2}), \ldots,
  r_m(\overline{Y_m}), G_1)\theta$; 
\item[MSW:] $G = \mathtt{msw}(\rv(\overline{X}), Y), G_1$: then
  $G' = G_1$;
\item[CONS:]$G = \id{Constr}, G_1$ and $\id{Constr}$ is satisfiable: then $G' = G_1$.
\end{description}
A \emph{symbolic derivation} of $G$ is a sequence of goals $G_0, G_1,
\ldots$ such that $G=G_0$ and, for all $i \geq 0$, $G_i \rightarrow G_{i+1}$.
\end{Def}
Note that the traditional notion of derivation in a logic program
coincides with that of symbolic derivation when the selected subgoal
(literal) is not an \texttt{msw} or a constraint. When the selected
subgoal is an \texttt{msw}, PRISM's inference will construct the next
step by enumerating the values of the random variable.  In contrast,
symbolic derivation skips \texttt{msw}'s and constraints and
continues with the remaining subgoals in a goal.  The effect of these
constructs is computed by associating (a) variable type information
and (b) a success function (defined below) with each goal in the
derivation.  The symbolic derivation for the goal \texttt{fmix(X)} over the program
in Example~\ref{ex:fmm} is shown in
Fig.~\ref{fig:fmmderivation}.

\paragraph{Success Functions:}
Goals in a symbolic derivation may contain variables whose values are
determined by \texttt{msw}'s appearing subsequently in the
derivation.  With each goal $G_i$ in a symbolic derivation, we
associate a set of variables, $V(G_i)$, that is a subset of variables
in $G_i$.  The set $V(G_i)$ is such that the variables in $V(G_i)$ 
subsequently appear as parameters or
outcomes of \texttt{msw}'s in some subsequent  goal $G_j$, $j \geq i$.
We can further partition $V$ into two disjoint sets, $V_c$ and $V_d$,
representing continuous and discrete variables,
respectively.  

Given a goal $G_i$ in a symbolic derivation, we can associate with it a
\emph{success function}, which is a function from the set of all
valuations of $V(G_i)$ to $[0,1]$.  Intuitively, the success function
represents the probability that the symbolic derivation represents a
successful derivation for each valuation of $V(G_i)$. 
Note that the success function computation uses a set of distribution parameters $\Theta$.
For simplicity, we often omit it in the equations and use it when it's not clear from the context.
 
\paragraph{Representation of success functions:}
Given a set of variables $\mathbf{V}$, let $\mathbf{C}$ denote the set
of all linear equality constraints over reals using
$\mathbf{V}$.  Let $\mathbf{L}$ be the set of all linear functions over
$\mathbf{V}$ with real coefficients.  Let
$\mathcal{N}_{X}(\mu,\sigma^2)$ be the PDF of a univariate Gaussian
distribution with mean $\mu$ and variance $\sigma^2$, and $\delta_{x}(X)$ be the Dirac delta function which is zero everywhere except at $x$ and integration of the delta function over its entire range is 1.  Expressions of
the form $k*\prod_{l} \delta_{v}(V_{l}) \prod_{i} \mathcal{N}_{f_i}$, where $k$ is a non-negative
real number and $f_i \in \mathbf{L}$, are called \emph{product PDF
  (PPDF) functions 
  over} $\mathbf{V}$.  We use $\phi$ (possibly subscripted) to denote
such functions.  A pair $\langle \phi, C\rangle$ where $C \subseteq \mathbf{C}$ is
called a \emph{constrained PPDF function}. A sum of a finite number of
constrained PPDF functions is called a \emph{success function}, represented as
$\sum_{i} \langle \phi_{i}, C_{i} \rangle$.

We use $C_{i}(\psi)$ to denote the
constraints (i.e., $C_{i}$) in the $i^{th}$ constrained PPDF function of
success function $\psi$;
and $D_{i}(\psi)$ to denote the $i^{th}$ PPDF
function of $\psi$.

\paragraph{Success functions of base predicates:}
The success function of a constraint $C$ is $\langle 1, C\rangle$.
The success function of \emph{true} is $\langle 1, \id{true}\rangle$.
The PPDF component of $\mathtt{msw}(\rv(\overline{X}), Y)$'s success
function  is the
probability density function of $\rv$'s distribution if $\rv$ is
continuous, and its probability mass function if $\rv$ is discrete; its constraint component
is \id{true}.

\begin{Ex}
The success function of \texttt{msw(m,M)} for the program in
Example~\ref{ex:fmm} is $\psi_1 = 0.3 \delta_{a}(M) + 0.7 \delta_{b}(M)$.

The success function \texttt{msw(w(M), X)} for the program in
Example~\ref{ex:fmm} is 
$\psi_2 =  \delta_{a}(M) \mathcal{N}_{X}(2.0, 1.0) +  \delta_{b}(M) \mathcal{N}_{X}(3.0, 1.0)$.
\qed
\end{Ex}

\paragraph{Success functions of user-defined predicates:}
If $G \rightarrow G'$ is a step in a derivation, then the success
function of $G$ is computed bottom-up based on the success
function of $G'$.  This computation is done using \emph{join} and
\emph{marginalize} operations on success functions.

\begin{Def} [Join]
Let $\psi_{1}=\sum_{i} \langle D_{i}, C_{i} \rangle$ and $\psi_{2} =
\sum_{j} \langle D_{j}, C_{j} \rangle$ be two success functions, then
join of $\psi_{1}$ and $\psi_{2}$ represented as $\psi_{1} * \psi_{2}$
is the success function $\sum_{i,j} \langle D_{i}D_{j}, C_{i} \wedge
C_{j} \rangle$. 
\end{Def}

Given a success function $\psi$ for a goal $G$, the success function
for $\exists X.\ G$ is computed by the marginalization operation.
Marginalization w.r.t. a discrete variable is straightforward and
omitted.  Below we
define marginalization w.r.t. continuous variables in two steps: first
rewriting the success function in a projected form and then doing the
required integration.

Projection eliminates any linear constraint on $V$, where $V$ is the continuous variable to marginalize over. The projection operation, denoted by $\psi\downarrow_{V}$, involves finding a linear constraint 
(i.e., $V = \overline{a} \cdot \overline{X}+b$) on $V$ and replacing all occurrences of $V$ in the success function by  $\overline{a} \cdot \overline{X}+b$.

\begin{Pro}
\label{integration}
Integration of a PPDF function with respect to a variable $V$ is a PPDF function, i.e., 
\begin{align*}
 \alpha \int^{\infty}_{-\infty} \prod_{k=1}^{m}
\mathcal{N}_{(\overline{a_{k}} \cdot
  \overline{X_{k}}+b_{k})}(\mu_{k},\sigma^{2}_{k}) dV \\= \alpha' \prod_{l=1}^{m'}
\mathcal{N}_{(\overline{a'_{l}} \cdot
  \overline{X'_{l}})+b'_{l}}(\mu'_{l},\sigma'^{2}_{l})
\end{align*}
where $V \in \overline{X_{k}}$ and $V \not\in \overline{X'_{l}}$.\\
\end{Pro}

\begin{Def} [Integration]
Let $\psi$ be a success function that does not contain any linear constraints on $V$. Then integration of $\psi$ with
 respect to $V$, denoted by $\oint_{V}\psi$ is a
success function $\psi'$ such that
$\forall i. D_{i}(\psi') = \int D_{i}(\psi) dV$.
\end{Def}

\begin{Def} [Marginalize]
Marginalization of a success function $\psi$ with respect to a
variable $V$, denoted by $\marginalize(\psi, V)$, is a success
function $\psi'$ such that
\begin{align*}
\psi'  &= \oint_{V} \psi \downarrow_{V} 
\end{align*}
\end{Def}

We overload $\marginalize$ to denote marginalization over a set of
variables, defined such that
$\marginalize(\psi, \{V\} \cup \overline{X}) =
\marginalize(\marginalize(\psi, V), \overline{X})$ and
$\marginalize(\psi, \{\}) = \psi$.

The success function for a derivation is defined as
follows.
\begin{Def}[Success function of a derivation]
 Let $G \rightarrow G'$.  Then the success function of $G$, denoted by
 $\psi_G$, is computed from that of $G'$, based on the way $G'$ was derived:
\begin{description}
\item[PCR:] $\psi_G = \marginalize(\psi_{G'}, V(G') - V(G))$.
\item[MSW:] Let $G= \kw{msw}(\rv(\overline{X}), Y), G_1$.
  Then $\psi_G = \psi_{msw(rv(\overline{X}),Y)} * \psi_{G'}$.
\item[CONS:] Let $G= Constr, G_1$.  Then
  $\psi_G = \psi_{Constr} * \psi_{G'}.$
\end{description}
\end{Def}
Note that the above definition carries PRISM's assumption that an
instance of a random variable occurs at most once in any derivation.
In particular, the PCR step marginalizes success functions w.r.t. a
set of variables; the valuations of the set of variables must be 
mutually exclusive for correctness of this step.  The MSW step joins
success functions; the goals joined must use independent random
variables for the join operation to correctly compute success
functions in this step.

\begin{Ex}
  Fig.~\ref{fig:fmmderivation} shows the symbolic derivation
  for goal \texttt{fmix(X)} over the finite mixture model program in
  Example~\ref{ex:fmm}.  Success function of goal $G_{3}$ is
  $\psi_{msw(w(M),X)}(M,X)$, hence $\psi_{G_3} = \delta_{a}(M)
  \mathcal{N}_{X}(\mu_{a}, \sigma^{2}_{a}) + \delta_{b}(M) \mathcal{N}_{X}(\mu_{b}, \sigma^{2}_{b})$.

  $\psi_{G_{2}}$ is $\psi_{msw(m, M)}(M) * \psi_{G_{3}}(M,X)$ which yields $\psi_{G_2} = p_{a} \delta_{a}(M)
   \mathcal{N}_{X}(\mu_{a}, \sigma^{2}_{a}) +  p_{b} \delta_{b}(M) \mathcal{N}_{X}(\mu_{b}, \sigma^{2}_{b})$.
Note that  $\delta_{b}(M)\delta_{a}(M)=0$ as $M$ can not be both $a$ and $b$ at the same time.
Also $\delta_{a}(M)\delta_{a}(M) = \delta_{a}(M)$.

  Finally, $\psi_{G_1} = \marginalize(\psi_{G_2}, M)$ which is $p_{a} 
  \mathcal{N}_{X}(\mu_{a}, \sigma^{2}_{a}) + p_{b}  \mathcal{N}_{X}(\mu_{b}, \sigma^{2}_{b})$.  Note
  that $\psi_{G_1}$ represents the mixture distribution ~\cite{fmm} of
  mixture of two Gaussian distributions.  
  
  Here $p_{a} = 0.3, p_{b} = 0.7, \mu_{a} = 2.0, \mu_{b} = 3.0$, and $\sigma^{2}_{a} = \sigma^{2}_{b} = 1.0$.
  \qed
\end{Ex}

Note that for a program with only discrete random variables, there may be
exponentially fewer symbolic derivations than concrete derivations
\emph{a la} PRISM.  The compactness is only in terms of \emph{number}
of derivations and not the total size of the representations.  In
fact, for programs with only discrete random variables, there is a
one-to-one correspondence between the entries in the tabular
representation of success functions and PRISM's answer tables.  For
such programs, it is easy to show that the time complexity of the
inference presented in this paper is same as that of PRISM.

%%% Local Variables:
%%% mode: latex
%%% TeX-master: "main"
%%% End:

\section{Learning}
\label{sec:Learning}

We use the expectation-maximization algorithm~\cite{DempsterEM} to learn the distribution parameters from data. First we show how to compute the expected sufficient statistics (ESS) of the random variables and then describe our algorithm.

The ESS of a discrete random variable is a n-tuple where $n$ is the
number of values that the discrete variable takes. Suppose that a
discrete random variable $V$ takes $v_{1}, v_{2}, ..., v_{n}$ as
values. Then the ESS of $V$ is $(ESS^{V=v_{1}}, ESS^{V=v_{2}}, ...,
ESS^{V=v_{n}})$ where $ESS^{V=v_{i}}$ is the expected number of times
variable $V$ had valuation $v_i$ in all possible proofs for a goal.  
The ESS of a Gaussian random variable $X$ is a triple $(ESS^{X,\mu},
ESS^{X,\sigma^{2}}, ESS^{X,count})$ where the components denote the
expected sum, expected sum of squares and the expected number of uses
of random variable $X$, respectively, in all possible proofs of a
goal.  When derivations are enumerated, the ESS for each random
variable can be represented by a tuple of reals.  To accommodate
\emph{symbolic} derivations, we lift each component of ESS to a
function, represented as described below.

\paragraph{Representation of ESS functions:}
For each component $\nu$ (discrete variable valuation, mean, variance, total counts) of a random variable, its ESS function in a goal $G$ is represented as follows:
\[
\xi^{\nu}_{G} = \sum_{i} \langle \chi_{i} \phi_{i}, C_{i} \rangle.
\]
where $\langle \phi_{i}, C_{i} \rangle$ is a constrained PPDF function and 
\[
\chi_{i}  = \left\{
\begin{array}{ll}
\overline{a}_{i} \cdot \overline{X}_{i}+b_{i} & \text{if $\nu$ = $X,\mu$}\\
\overline{a}_{i} \cdot \overline{X}_{i}^{2}+b_{i} & \text{if $\nu$ =$X,\sigma^2$}\\
b_{i}  & \text{otherwise}\\
\end{array}\right.
\]

\noindent
Here $\overline{a}_{i}, b_{i}$ are constants, and $\overline{X}_{i} = V_{c}(G)$. 

Note that the representation of ESS function is same as that of
success function for discrete random variable valuations and total
counts. \emph{Join} and \emph{Marginalize} operations, defined earlier
for success functions, can be readily defined for ESS functions as
well.  The computation of ESS functions for a goal, based on the
symbolic derivation, uses the extended \emph{join} and
\emph{marginalize} operations.  The set of all ESS functions is closed
under the extended \emph{Join} and \texttt{Marginalize} operations.

\paragraph{ESS functions of base predicates:}
The ESS function of the $i^{th}$ parameter of a discrete random variable $V$ is $P(V=v_i)\delta_{v_{i}}(V)$.
The ESS function of the mean of a continuous random variable $X$ is $X
\mathcal{N}_{X}(\mu, \sigma^{2})$, and the ESS function of the
variance of a continuous random variable $X$ is $X^{2}
\mathcal{N}_{X}(\mu, \sigma^{2})$. Finally, the ESS function of the
total count of a continuous random variable $X$ is
$\mathcal{N}_{X}(\mu, \sigma^{2})$. 

\begin{Ex}
In this example, we compute the ESS functions of the random variables (\texttt{m}, \texttt{w(a)}, and \texttt{w(b)}) in Example~\ref{ex:fmm}.
According to the definition of ESS function of base predicates, the ESS functions of these random variables for goals $msw(m, M)$ and  $msw(w(M), X)$ are
 \begin{center}
\begin{tabular}{|c|c|c|c|}
  \hline
  ESS & for $msw(m, M)$ & for $msw(w(M), X)$ \\ \hline
  $\xi^{k}$ & $p_{k} \delta_{k}(M)$ & $0$  \\ \hline
  $\xi^{\mu_{k}}$ & $0$ & $X \delta_{k}(M) \mathcal{N}_{X}(\mu_{k}, \sigma^{2}_{k})$  \\ \hline
  $\xi^{\sigma^{2}_{k}}$ & $0$ & $X^{2} \delta_{k}(M) \mathcal{N}_{X}(\mu_{k}, \sigma^{2}_{k})$  \\ \hline
  $\xi^{count_{k}}$ & $0$ & $\delta_{k}(M) \mathcal{N}_{X}(\mu_{k}, \sigma^{2}_{k})$  \\ \hline
\end{tabular}
\end{center}
where $k \in \{a,b\}$.
\qed
\end{Ex}

\paragraph{ESS functions of user-defined predicates:}
If $G \rightarrow G'$ is a step in a derivation, then the ESS 
function of a random variable for $G$ is computed bottom-up based on
the its ESS
function for $G'$. 

The ESS function of a random variable component in a derivation is defined as
follows.
\begin{Def}[ESS functions in a derivation]
 Let $G \rightarrow G'$.  Then the ESS function of a random variable component $\nu$ in the goal $G$, denoted by
 $\xi^{\nu}_G$, is computed from that of $G'$, based on the way $G'$ was derived:
\begin{description}
\item[PCR:] $\xi^{\nu}_G = \marginalize(\xi^{\nu}_{G'}, V(G') - V(G))$.
\item[MSW:] Let $G= \kw{msw}(\rv(\overline{X}), Y), G_1$.
  Then $\xi^{\nu}_G = \psi_{msw(rv(\overline{X}),Y)} * \xi^{\nu}_{G'} + \psi_{G'} * \xi^{\nu}_{msw}$.
\item[CONS:] Let $G= Constr, G_1$.  Then
  $\xi_G^{\nu} = \psi_{Constr} * \xi^{\nu}_{G'}.$
\end{description}
\end{Def}

\begin{Ex}
Using the definition of ESS function of a derivation involving MSW, we compute the ESS function 
of the random variables in goal $G_{2}$ of Fig.~\ref{fig:fmmderivation}.
 \begin{center}
\begin{tabular}{|c|c|c|}
  \hline
  &  ESS functions for goal $G_{2}$ \\ \hline
  $\xi^{k}$ & $ p_{k} \delta_{k}(M) \mathcal{N}_{X}(\mu_{k}, \sigma^{2}_{k})$  \\ \hline
  $\xi^{\mu_{k}}$ & $X p_{k} \delta_{k}(M) \mathcal{N}_{X}(\mu_{k}, \sigma^{2}_{k})$  \\ \hline
  $\xi^{\sigma^{2}_{k}}$ & $X^{2} p_{k} \delta_{k}(M) \mathcal{N}_{X}(\mu_{k}, \sigma^{2}_{k})$  \\ \hline
  $\xi^{count_{k}}$ & $p_{k} \delta_{k}(M) \mathcal{N}_{X}(\mu_{k}, \sigma^{2}_{k})$  \\ \hline
\end{tabular}
\end{center}
Notice the way $\xi^{k}_{G_{2}}$ is computed. 
\begin{align*}
\xi^{k}_{G_{2}} &= \psi_{msw(m, M)} \xi^{k}_{G_{3}} + \psi_{G_{3}} \xi^{k}_{msw(m,M)}\\
&= [p_{a} \delta_{a}(M) + p_{b} \delta_{b}(M)]. 0 \\&+  [\delta_{a}(M)
  \mathcal{N}_{X}(\mu_{a}, \sigma^{2}_{a}) + \delta_{b}(M) \mathcal{N}_{X}(\mu_{b}, \sigma^{2}_{b})].  p_{k} \delta_{k}(M)\\
&= p_{k} \delta_{k}(M) \mathcal{N}_{X}(\mu_{k}, \sigma^{2}_{k})
\end{align*}

Finally, for goal $G_{1}$ we marginalize the ESS functions w.r.t. $M$. 
\begin{center}
\begin{tabular}{|c|c|c|}
  \hline
  &  ESS functions for goal $G_{1}$ \\ \hline
  $\xi^{k}$ & $ p_{k} \mathcal{N}_{X}(\mu_{k}, \sigma^{2}_{k})$  \\ \hline
  $\xi^{\mu_{k}}$ & $X p_{k}  \mathcal{N}_{X}(\mu_{k}, \sigma^{2}_{k})$  \\ \hline
  $\xi^{\sigma^{2}_{k}}$ & $X^{2} p_{k}  \mathcal{N}_{X}(\mu_{k}, \sigma^{2}_{k})$  \\ \hline
  $\xi^{count_{k}}$ & $p_{k}  \mathcal{N}_{X}(\mu_{k}, \sigma^{2}_{k})$  \\ \hline
\end{tabular}
\end{center}\qed
\end{Ex}

The algorithm for learning distribution parameters ($\Theta$) uses a fixed set of training examples ($t_{1}, t_{2}, ..., t_{N}$). Note that the success and ESS functions for $t_{i}$'s are constants as the training examples are variable free (i.e., all the variables get marginalized over). 

\begin{Algo}[Expectation-Maximization]
\label{EMalgo}
Initialize the distribution parameters $\Theta$.
\begin{enumerate}
\item 
Construct the symbolic derivations for $\psi$ and $\xi$ using current $\Theta$.
\item
\textbf{E-step:} For each training example $t_{i}$ ($1\leq i \leq N$),
compute the ESS ($\xi_{t_{i}}$) of the random variables, and success
probabilities $\psi_{t_{i}}|$ w.r.t. $\Theta$. \\
\textbf{M-step:} Compute the MLE of the distribution parameters given the ESS and success probabilities (i.e., evaluate $\Theta'$).
$\Theta'$ contains updated distribution parameters ($p', \mu', \sigma'^{2}$).
More specifically, for a discrete random variable $V$, its parameters are updated as follows:
\begin{align*}
p'_{V= v}  =  \frac {\eta_{V=v}} {\sum_{u \in values(V)} \eta_{V=u}}
\end{align*}
where 
\begin{align*}
\eta_{V=v} = \sum_{i=1}^{N} \frac{\xi^{V=v}_{t_{i}}} {\psi_{t_{i}}}.
\end{align*}

For each continuous random variable $X$, its mean and variances are updated as follows: 
\begin{align*}
\mu'_{X}  =  \frac {\sum_{i=1}^{N} \frac{\xi^{X,\mu}_{t_{i}}} {\psi_{t_{i}}}} {N_{X}} 
\end{align*}
\begin{align*}
\sigma_{X}^{'2} &=  \frac {\sum_{i=1}^{N} \frac{\xi^{X,\sigma^{2}}_{t_{i}}  } {\psi_{t_{i}}}} {N_{X}} -  \mu_{X}^{'2} 
\end{align*}
where $N_{X}$ is the expected total count of $X$.
\begin{align*}
N_{X}  =  \sum_{i=1}^{N} \frac{\xi^{X,count}_{t_{i}}} {\psi_{t_{i}}}
\end{align*}
\item
Evaluate the log likelihood ($\ln P(t_{1},..,t_{N}|\Theta') = \sum_{i}
\ln \psi_{t_i}$) and check for convergence. Otherwise let $\Theta \leftarrow \Theta'$ and return to step 1. 
\end{enumerate}
\end{Algo}

\begin{Thm}
Algorithm~\ref{EMalgo} correctly computes the MLE which (locally) maximizes the likelihood.
\end{Thm}

\noindent (Proof) Sketch. The main routine of Algorithm~\ref{EMalgo}
for discrete case is same as the \emph{learn-naive} algorithm of \mycite{sato}, except the computation of $\eta_{V=v}$. 
\begin{align*}
\eta_{V=v} = \sum_{\text{for each goal } g} \frac{1} {\psi_{g}} \sum_{S} P(S)N^{v}_{S}.
\end{align*}
where $S$ is an explanation for goal $g$ and $N^{v}_{S}$ is the total number of times $V=v$ in $S$. 

We show that $\xi^{V=v}_{g} = \sum_{S} P(S)N^{v}_{S}$.

Let the goal $g$ has a single explanation $S$ where $S$ is a conjunction of subgoals (i.e., $S_{1:n} = g_{1}, g_{2}, ...,g_{n}$). Thus we need to show that $\xi^{V=v}_{g} = P(S)N^{v}_{S}$. 

We prove this by induction on the length of $S$. The definition of $\xi$ for base predicates gives the desired result for $n=1$. Let the above equation holds for length $n$ i.e., $\xi^{V=v}_{g_{1:n}} = P(S_{1:n})N_{1:n}$. For $S_{1:n+1} = g_{1}, g_{2}, ..., g_{n}, g_{n+1}$,
\begin{align*}
P(S_{1:n+1})N_{1:n+1} &= P(g_{1}, g_{2}, ..., g_{n}, g_{n+1})N_{1:n+1}\\
 &= \scriptstyle P(g_{1}, g_{2}, ..., g_{n}) P(g_{n+1}) (N_{1:n}+N_{n+1})\\
&= \scriptstyle P(S_{1:n}) P(g_{n+1}) N_{1:n} + P(S_{1:n}) P(g_{n+1}) N_{n+1}\\
&= \scriptstyle P(g_{n+1}) [P(S_{1:n})N_{1:n}] + P(S_{1:n}) [P(g_{n+1}) N_{n+1}]\\
&= P(g_{n+1}) \xi^{V=v}_{g_{1:n}}  + P(S_{1:n}) \xi^{V=v}_{g_{n+1}}\\
&= \xi^{V=v}_{g_{1:n+1}}
\end{align*}
The last step follows from the definition of $\xi$ in a derivation. 

Now based on the exclusiveness assumption, for disjunction (or multiple explanations) like $g = g_{1} \vee g_{2}$ it trivially follows that $\xi^{V=v}_{g} = \xi^{V=v}_{g_{1}} + \xi^{V=v}_{g_{2}}$.

\begin{Ex}
Let $x_{1}, x_{2}, ..., x_{N}$ be the observations. For a given training example $t_{i} = fmix(x_{i})$, the ESS functions are 
 \begin{center}
\begin{tabular}{|c|c|c|}
  \hline
  &  ESS functions for goal $fmix(x_{i})$ \\ \hline
  $\xi_{k}$ & $ p_{k} \mathcal{N}_{X}(x_{i}|\mu_{k}, \sigma^{2}_{k})$  \\ \hline
  $\xi_{\mu_{k}}$ & $x_{i} p_{k}  \mathcal{N}_{X}(x_{i}|\mu_{k}, \sigma^{2}_{k})$  \\ \hline
  $\xi_{\sigma^{2}_{k}}$ & $x_{i}^{2} p_{k}  \mathcal{N}_{X}(x_{i}|\mu_{k}, \sigma^{2}_{k})$  \\ \hline
  $\xi_{count_{k}}$ & $p_{k}  \mathcal{N}_{X}(x_{i}|\mu_{k}, \sigma^{2}_{k})$  \\ \hline
\end{tabular}
\end{center}
The E-step of the EM algorithm involves computation of the above ESS functions.

In the M-step, we update the model parameters from the computed ESS functions.
\begin{align}
\label{pa}
p'_{k} = \frac{\sum_{i=1}^{N} \frac{\xi^{k}_{t_i}} {\psi_{t_i}}} {\sum_{i=1}^{N} \frac{\xi^{a}_{t_i}} {\psi_{t_i}} + \frac{\xi^{b}_{t_i}} {\psi_{t_i}}} = \frac{\sum_{i=1}^{N} \frac{\xi^{k}_{t_i}} {\psi_{t_i}}} {N}
\end{align}
Similarly, 
\begin{align}
\label{mua}
\mu'_{k} &= \frac {\sum_{i=1}^{N} \frac{\xi^{\mu_{k}}_{t_i}} {\psi_{t_i}}} {N_{k}}
\end{align}
\begin{align}
\sigma_{k}^{'2}  &= \frac {\sum_{i=1}^{N} \frac{\xi^{\sigma_{k}^{2}}_{t_i}} {\psi_{t_i}}} {N_{k}} - \mu_{k}^{'2}
\end{align}
where $k \in \{a,b\}$ and
\begin{align}
N_{k} = \sum_{i=1}^{N} \frac{\xi^{count_k}_{t_i}} {\psi_{t_i}}
\end{align}
\end{Ex}

\begin{Ex}
This example illustrates that for the mixture model example, our ESS computation does the same computation as standard EM learning algorithm for mixture models~\cite{bishop}. 

Notice that for Equation~\ref{pa}, $\frac{\xi^{k}_{t_i}} {\psi_{t_i}} = \frac {p_{k}\mathcal{N}_{k}(x_{i}|\mu_{k},\sigma^{2}_{k}) } {\sum_{l} p_{l}\mathcal{N}_{l}(x_{i}|\mu_{l},\sigma^{2}_{l})}$ which is nothing but the posterior responsibilities presented in~\cite{bishop}. 
\begin{align*}
p'_{k} &= \frac{\sum_{i=1}^{N} \frac{\xi^{k}_{t_i}} {\psi_{t_i}}} {N}
&= \frac {\sum_{i=1}^{N} \frac{p_{k} \mathcal{N}_{X}(x_{i}|\mu_{k}, \sigma^{2}_{k})} {\sum_{l} p_{l}\mathcal{N}_{l}(x_{i}|\mu_{l},\sigma^{2}_{l})} } {N}
\end{align*}
Similarly for Equation~\ref{mua}, 
\begin{align*}
\mu'_{k} &= \frac {\sum_{i=1}^{N} \frac{\xi^{\mu_{k}}_{t_i}} {\psi_{t_i}}} {N_{k}}
&= \frac {\sum_{i=1}^{N} \frac{ x_{i} p_{k} \mathcal{N}_{X}(x_{i}|\mu_{k}, \sigma^{2}_{k})} {\sum_{l} p_{l}\mathcal{N}_{l}(x_{i}|\mu_{l},\sigma^{2}_{l})}}
{\sum_{i=1}^{N}  \frac{p_{k} \mathcal{N}_{X}(x_{i}|\mu_{k}, \sigma^{2}_{k})} {\sum_{l} p_{l}\mathcal{N}_{l}(x_{i}|\mu_{l},\sigma^{2}_{l})}}.
\end{align*}
Variances are updated similarly. 
\qed
\end{Ex}

%%% Local Variables:
%%% mode: latex
%%% TeX-master: "main"
%%% End:

\section{Discussion and Concluding Remarks}
\label{sec:extn}

The symbolic inference and learning procedures enable us to reason over 
a large class of statistical models such as hybrid Bayesian networks with discrete child-discrete parent, continuous child-discrete parent (finite mixture model), and continuous child-continuous parent (Kalman filter), which was hitherto not possible in PLP frameworks.  It can also be used for  hybrid models, e.g., models that mix discrete and Gaussian distributions.  For instance, consider the mixture model example (Example~\ref{ex:fmm}) where \texttt{w(a)} is Gaussian but \texttt{w(b)} is a discrete distribution with values $1$ and $2$ with $0.5$ probability each. The density of the mixture distribution can be written as 
\begin{align*}
f(X) = 0.3 \mathcal{N}_{X}(2.0, 1.0)  + 0.35  \delta_{1.0}(X)  + 0.35  \delta_{2.0}(X)
\end{align*}
Thus the language can be used to model problems that lie outside traditional hybrid Bayesian networks.

ProbLog and LPAD do not impose PRISM's mutual exclusion and independence restrictions.  Their inference technique first materializes the set of explanations for each query, and represents this set as a BDD, where each node in the BDD is a (discrete) random variable.  Distinct paths in the BDD are mutually exclusive and variables in a single path are all independent. Probabilities of query answers are computed trivially based on this BDD representation.  The technical development in this paper is limited to PRISM and imposes its restrictions.  However, by materializing the set of symbolic derivations first, representing them in a factored form (such as a BDD) and then computing success functions on this representation, we can readily lift the restrictions for the parameter learning technique.

This paper considered only univariate
Gaussian distributions.  Traditional parameter learning techniques have been described for multivariate distributions without introducing additional machinery.  Extending our learning algorithm to the multivariate case is a topic of future work.

\bibliographystyle{aaai}
\bibliography{biblio}

\end{document}